\documentclass[runningheads]{llncs}
\usepackage[T1]{fontenc}
\usepackage{graphicx,verbatim}
\usepackage{amsmath}
\usepackage{amssymb}
\usepackage{booktabs}
\usepackage{multirow}
\usepackage{threeparttable}
\usepackage{graphicx}
\usepackage[table]{xcolor}
\begin{document}
\title{SAFE-Diff: Scale-Aware Attention and Feature-Dispersive Diffusion with Uncertainty Estimation for Contrast-Enhanced Breast MRI Synthesis}
\titlerunning{SAFE-Diff}
%



\author{Tianyu Zhang\inst{1,2,3,\dag,*} 
\and Xinglong Liang\inst{1,2,\dag} 
\and Jarek van Dijk\inst{1,2,\dag} 
\and Luyi Han\inst{1,2,4} 
\and Chunyao Lu\inst{1,2} 
\and Antonio Portaluri\inst{1,2} 
\and Xinghe Xie\inst{4} 
\and Yaofei Duan\inst{1,2,4} 
\and Nika Rasoolzadeh\inst{1,2} 
\and Xin Wang\inst{2} 
\and Yuan Gao\inst{2} 
\and Muzhen He\inst{2} 
\and Yue Sun\inst{4} 
\and Jonas Teuwen\inst{5} 
\and Tao Tan\inst{4} 
\and Ritse Mann\inst{1,2,*} 
}

\authorrunning{T. Zhang et al.}
\institute{Department of Medical Imaging, Radboud University Medical Center, Geert Grooteplein 10, 6525 GA, Nijmegen, The Netherlands \and Department of Radiology, Netherlands Cancer Institute, Plesmanlaan 121, 1066 CX, Amsterdam, The Netherlands \and Maastro Clinic, Dr. Tanslaan 12, 6229 ET, Maastricht, The Netherlands \and Faculty of Applied Science, Macao Polytechnic University, 999078, Macao, China \and Department of Radiation Oncology, Netherlands Cancer Institute, Plesmanlaan 121, 1066 CX, Amsterdam, The Netherlands  \\
$\dag$ T. Z., X. L. and J. D. contributed equally to this work.\\
* Correspondence: \email{Tianyu.Zhang@radboudumc.nl; Ritse.Mann@radboudumc.nl}}
  
\maketitle              
\begin{abstract}

Synthesizing high-fidelity contrast-enhanced MRI from non-contrast sequences is clinically valuable for safer and more efficient breast cancer screening, yet remains challenging due to complex lesion textures and heterogeneous enhancement patterns. In this work, we propose SAFE-Diff, a multi-scale attention–enhanced diffusion network for contrast-agent-free breast MRI synthesis. The framework combines convolutional feature extraction with complementary attention mechanisms across spatial scales, while a window-based self-attention module inspired by Swin Transformers is employed at the bottleneck to capture long-range anatomical context via shifted-window interactions. To improve robustness in heterogeneous tissues, we introduce an uncertainty-aware learning formulation that adaptively reweights the reconstruction objective based on spatially varying prediction confidence. In addition, a feature-dispersive learning strategy is applied to intermediate latent representations to mitigate representation redundancy and preserve fine-grained diagnostic details. Extensive experiments demonstrate that SAFE-Diff accurately reconstructs contrast-enhancement characteristics and outperforms state-of-the-art approaches. Code is available at \url{https://github.com/Netherlands-Cancer-Institute/SAFE-Diff}.


\end{abstract}
\section{Introduction}
Breast cancer remains one of the most prevalent malignancies and a leading cause of cancer-related mortality among women worldwide ~\cite{bray2024global,lu2025ai,zhang2023radiomics}. Dynamic Contrast-Enhanced MRI (DCE-MRI) is the most sensitive imaging modality for breast cancer detection and characterization~\cite{mann2019breast}. By capturing contrast-agent uptake and washout kinetics, DCE-MRI provides essential information on tumor vascularity and permeability, enabling early diagnosis and treatment planning.

However, its reliance on gadolinium-based contrast agents restricts widespread use in large-scale screening. Contrast administration increases scan time, cost, and clinical complexity, while safety concerns—including gadolinium deposition, nephrogenic systemic fibrosis, and allergic reactions—have raised significant clinical and environmental considerations~\cite{nguyen2020dentate,marckmann2006nephrogenic,olchowy2017presence}. These limitations underscore the urgent need for a “virtual DCE” alternative that maintains diagnostic performance without exogenous contrast agents.

Among non-contrast breast MRI sequences, pre-contrast T1-weighted imaging and diffusion-weighted imaging (DWI) are particularly attractive candidates for enabling virtual contrast enhancement. DWI is routinely available in clinical protocols without requiring contrast administration, improving feasibility when contrast use is undesirable or impractical. Moreover, DWI provides complementary functional sensitivity to tissue cellularity and microstructural heterogeneity, which are relevant for lesion detection and characterization~\cite{baltzer2020diffusion,van2021factors}. Combined with the stable anatomical context of pre-contrast T1, diffusion-informed cues can help emphasize lesion--parenchyma differences and tumor-related tissue properties, motivating their use as practical conditioning signals for synthesizing enhancement-like appearance.

Recent studies have explored Generative Adversarial Networks (GAN) for cross-modality medical image synthesis, including the generation of contrast-enhanced MRI from non-contrast acquisitions, demonstrating promising visual realism and structural consistency~\cite{chung2022deep,han2024synthesis,han2023explainable,muller2023using,zhang2023synthesis,zhang2024important}. However, GAN-based approaches can be sensitive to adversarial training dynamics and may trade strict fine-grained structural fidelity for perceptual realism, which is an important consideration in oncologic imaging. Diffusion models, by contrast, provide a more stable likelihood-based generative framework with improved distribution coverage and progressive detail refinement, making them particularly suitable for high-fidelity contrast-enhanced MRI synthesis ~\cite{jiang2023cola,dayarathna2025mu,pan20232d}.

In this work, we propose SAFE-Diff (\textbf{S}cale-aware \textbf{A}ttention and \textbf{F}eature-dispersive framework with \textbf{E}stimated Uncertainty \textbf{Diff}usion), a multi-scale attention diffusion network for robust contrast-enhanced breast MRI synthesis, integrating uncertainty-aware modeling and feature-dispersive learning. The main contributions are summarized as follows: (1) A direct $x_0$-prediction diffusion framework with multi-scale attention for hierarchical anatomical modeling. (2) A heteroscedastic uncertainty-aware formulation that jointly predicts the synthesized image and pixel-wise variance for adaptive reconstruction weighting. (3) A feature-dispersive regularization on intermediate latent representations to reduce redundancy and preserve fine-grained structural details.

\section{Materials and Methods}
\subsection{Problem Formulation}
Let $x \in \mathbb{R}^{H \times W}$ denote a target contrast-enhanced breast MRI slice, and $\mathbf{c} \in \mathbb{R}^{H \times W \times C}$ denote the corresponding non-contrast conditioning inputs, including T1-weighted and diffusion-weighted images.

Our objective is to synthesize a realistic contrast-enhanced MRI $\hat{x}_0$ conditioned on $\mathbf{c}$, while explicitly modeling spatially varying uncertainty and preserving fine-grained anatomical and textural details critical for clinical interpretation.

\begin{figure}[htbp]
    \centering
    \includegraphics[width=\textwidth, keepaspectratio]{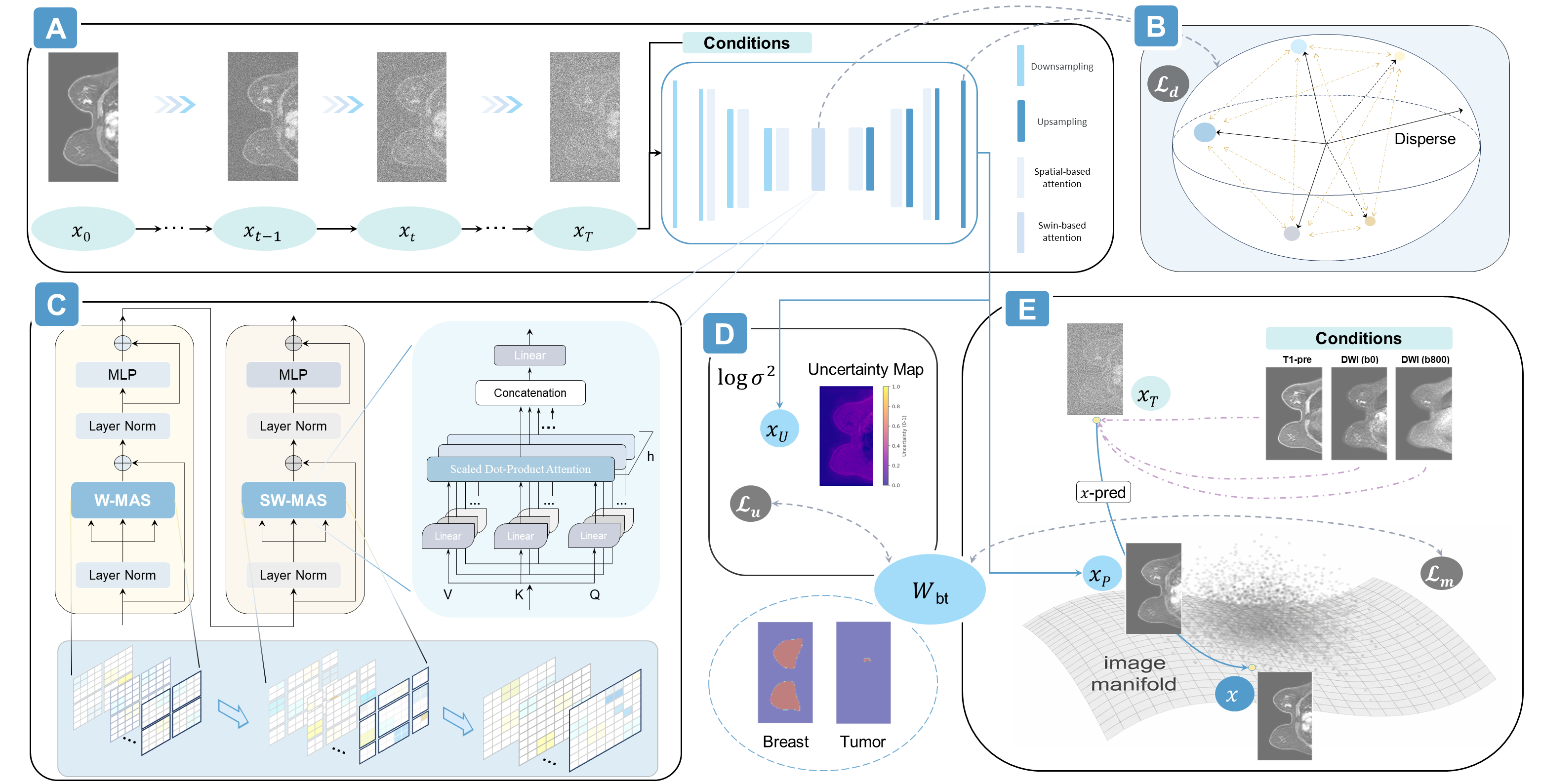}
    \caption{The flowchart of this study.}
    \label{fig1}
\end{figure}
\subsection{Direct $x_0$-Prediction Diffusion Framework}

SAFE-Diff adopts a diffusion-based generative framework that directly predicts the clean image $x_0$ rather than the noise residual.
Given a diffusion timestep $t$, Gaussian noise is added to the target image:
\begin{equation}
x_t = x_0 + \boldsymbol{\epsilon}_t, 
\quad \boldsymbol{\epsilon}_t \sim \mathcal{N}(0, \sigma_t^2).
\end{equation}
The network regresses $x_0$ from noisy input $x_t$ and condition $\mathbf{c}$:
\begin{equation}
(\boldsymbol{\mu}, \log \boldsymbol{\sigma}^2) = f_\theta([x_t, \mathbf{c}], t),
\end{equation}
where $f_\theta(\cdot)$ denotes the proposed Swin Transformer–based diffusion network, and the output consists of $\boldsymbol{\mu}$, the predicted clean image $\hat{x}_0$, and $\log \boldsymbol{\sigma}^2$,  a pixel-wise uncertainty estimate. Fig.~\ref{fig1} shows the flowchart of this study. Compared with noise-prediction objectives, direct $x_0$ regression simplifies the learning target and provides more stable convergence for high-resolution medical images (as shown in Fig.~\ref{fig1}E). 

\subsection{Multi-Scale Attention Architecture}
The SAFE-Diff architecture follows a U-shaped diffusion backbone integrating convolutional feature extraction and multi-scale attention mechanisms.

\paragraph{Encoder–Decoder Structure}
The encoder progressively downsamples the input through multiple stages composed of convolutional blocks followed by optional spatial attention modules. Let $\mathbf{h}^{(l)} \in \mathbb{R}^{C_l \times H_l \times W_l}$ denote the feature map at scale $l$.  
Each downsampling stage is defined as:

\begin{equation}
\mathbf{h}^{(l+1)} = \mathcal{D}\big( \mathcal{A}_s ( \mathcal{C}(\mathbf{h}^{(l)}) ) \big),
\end{equation}
where $\mathcal{C}(\cdot)$ denotes convolutional feature extraction, $\mathcal{A}_s(\cdot)$ represents spatial attention (when enabled), and $\mathcal{D}(\cdot)$ denotes spatial downsampling. The decoder mirrors the encoder structure via upsampling layers and skip connections to fuse multi-scale representations.

\paragraph{Spatial Attention at Intermediate Scales}
At intermediate resolutions, we employ a spatial self-attention mechanism to enhance salient local responses. Given feature map $\mathbf{h} \in \mathbb{R}^{C \times H \times W}$, spatial attention is computed as:

\begin{equation}
\mathbf{Q} = W_q \mathbf{h}, \quad
\mathbf{K} = W_k \mathbf{h}, \quad
\mathbf{V} = W_v \mathbf{h},
\end{equation}
where $W_q$, $W_k$, $W_v$ are 1×1 convolutional projections.
The attention map is then obtained by:
\begin{equation}
\mathbf{A} = \text{softmax}\left(\frac{\mathbf{Q}^\top \mathbf{K}}{\sqrt{C}}\right),
\end{equation}
and the output is given by:
\begin{equation}
\hat{\mathbf{h}} = \mathbf{h} + W_0 (\mathbf{V} \mathbf{A}),
\end{equation}
where $W_0$ is a projection layer.
This formulation enables adaptive emphasis of spatially salient structures while preserving local texture continuity.

\paragraph{Window-Based Self-Attention at the Bottleneck}

At the lowest spatial resolution, we introduce a window-based self-attention mechanism inspired by Swin Transformers to model long-range anatomical dependencies efficiently.
The bottleneck feature map is partitioned into non-overlapping windows, within which self-attention is computed:

\begin{equation}
\text{W-MSA}(\mathbf{z}) =
\text{softmax}\left(\frac{\mathbf{Q}\mathbf{K}^\top}{\sqrt{d}}\right)\mathbf{V},
\end{equation}
where $\mathbf{z}$ denotes windowed features and $d$ is the feature dimension.

To enable cross-window information exchange, shifted-window self-attention (SW-MSA) is alternated across successive layers:

\begin{equation}
\mathbf{z}^{(l+1)} =
\begin{cases}
\text{W-MSA}(\mathbf{z}^{(l)}), & l \text{ even}, \\
\text{SW-MSA}(\mathbf{z}^{(l)}), & l \text{ odd}.
\end{cases}
\end{equation}

\subsection{Uncertainty-Aware Learning}
To model spatially varying ambiguity, the network jointly predicts the synthesized image $\boldsymbol{\mu}$ and a pixel-wise log-variance map $\log \boldsymbol{\sigma}^2$ (Fig.~\ref{fig1}D). 
The uncertainty-aware reconstruction objective adaptively weights pixel-wise errors, assigning higher confidence to reliable regions while attenuating gradients in ambiguous areas. 
This formulation enhances robustness in challenging regions such as tumor boundaries and heterogeneous parenchyma. Uncertainty-Aware Reconstruction Loss:

\begin{equation}
\mathcal{L}_{\text{unc}} =
\frac{1}{N} \sum_{i=1}^{N}
\omega_i
\left[
\exp(-\log \sigma_i^2)
(\mu_i - x_i)^2
+ \log \sigma_i^2
\right],
\end{equation}
where $N$ denotes the number of pixels and $\omega_i$ a spatial weighting factor.

\subsection{Feature-Dispersive Learning Strategy}
To reduce representation redundancy in deep transformer layers, a feature-dispersive objective is applied to intermediate latent features from selected stages (Fig.~\ref{fig1}B). The loss penalizes similarity between normalized feature embeddings within a batch, promoting diverse and expressive representations. It is imposed at the bottleneck transformer layer and the final decoder stage.
Given batch features $\{ \mathbf{f}_k \}_{k=1}^B$, L2-normalized embeddings are defined as:

\begin{equation}
\tilde{\mathbf{f}}_k =
\frac{\mathbf{f}_k}{\|\mathbf{f}_k\|_2},
\end{equation}

\begin{equation}
\mathcal{L}_{\text{disp}} =
\frac{1}{B}
\sum_{k=1}^{B}
\log
\sum_{j \neq k}
\exp
\left(
\frac{\tilde{\mathbf{f}}_k^\top \tilde{\mathbf{f}}_j}{\tau}
\right).
\end{equation}

\subsection{Mask-Aware Multi-Scale Perceptual Consistency Loss}
To preserve structural and textural fidelity beyond pixel-wise similarity, we employ a mask-aware perceptual loss computed using a pretrained VGG-16 network. Unlike conventional perceptual losses that treat all spatial locations equally~\cite{johnson2016perceptual,zhang2018unreasonable}, our formulation integrates spatial weighting derived from anatomical masks to emphasize clinically relevant regions. Mask-Aware Multi-Scale Perceptual Loss:

\begin{equation}
\mathcal{L}_{\text{perc}} =
\sum_{l \in \mathcal{S}}
\lambda_l
\left\|
W_l \odot
\left(
\phi_l(\boldsymbol{\mu})
-
\phi_l(x)
\right)
\right\|_1,
\end{equation}
where $W_l$ denotes the scale-aligned spatial weight map, $\odot$ represents element-wise multiplication, $\mathcal{S}$ denotes a set of selected layers spanning shallow to deep representations, $\lambda_l$ controls the contribution of each feature scale.

\subsection{Training Objective}

All losses are computed on $\boldsymbol{\mu}$ corresponding to direct $x_0$ regression. The final training objective is expressed as:

\begin{equation}
\mathcal{L}_{\text{total}} =
\mathcal{L}_{\text{unc}}
+
\alpha \mathcal{L}_{\text{perc}}
+
\beta \mathcal{L}_{\text{disp}},
\end{equation}
where $\alpha$ balances perceptual consistency, $\beta$ controls the contribution of feature-dispersive learning.


\subsection{Data Collection}

This retrospective study was approved by the institutional review board with waived informed consent. Breast MRI examinations acquired between January 2000 and March 2025 at an in-house center were reviewed. After quality control excluding severe motion artifacts, missing sequences, or incomplete diagnostic results, 3,308 patients were retained. For each patient, pre-contrast T1-weighted (T1-pre), post-contrast T1-weighted (T1-post, ground truth), and DWI at $b=0$ and $b=800$ s/mm$^2$ were collected. All sequences were spatially aligned and processed into 2D slices with consistent in-plane resolution. The in-house cohort was split into training, validation, and testing sets (70\%/20\%/10\%) at the patient level. An external cohort of 242 patients from the I-SPY2 trial was used exclusively for cross-institutional evaluation~\cite{newitt2025acrin}.

\subsection{Experimental Setup and Implementation Details}

The model was implemented in PyTorch using a direct $x_0$-prediction diffusion framework with 1{,}000 training timesteps and 15 reverse steps at inference (Heun scheduler). A U-shaped backbone integrated convolutional encoding--decoding and multi-scale attention, with feature-dispersive regularization applied to selected decoder layers. The model was trained for 100 epochs.

The perceptual weight $\alpha$ was progressively increased (1.0 for epochs 0--4, 5.0 for 5--9, 10.0 for 10--19, and 20.0 thereafter), while the feature-dispersive weight was fixed at $\beta=0.002$ with temperature $\tau=0.1$. Spatial weights in the uncertainty-aware reconstruction were set to 1.0 (background), 20.0 (breast), and 1000.0 (tumor), followed by batch-wise mean normalization. Log-variance predictions were clamped to $[-1.5,\,3.0]$ for stability. Perceptual loss was computed using VGG-16 features (relu1\_2--relu5\_3) with layer weights [0.25, 0.5, 0.5, 1.0, 1.0]. Optimization used AdamW ($1\times10^{-4}$, weight decay $1\times10^{-2}$) with two-epoch warm-up (0.1$\times$), gradient clipping (1.0), mixed-precision training, and distributed data parallelism. Comparisons were conducted against MT-DDPM~\cite{pan20232d}, Cola-Diff~\cite{jiang2023cola}, and MU-Diff~\cite{dayarathna2025mu}.

Ablation experiments on the in-house validation set incrementally added uncertainty-aware learning, feature-dispersive regularization, and perceptual loss to a reconstruction baseline. Performance was evaluated under global and tumor-region settings using Structural Similarity Index Measure (SSIM), Peak Signal-to-Noise Ratio (PSNR), Normalized Mean Squared Error (NMSE), and normalized High-Frequency Error Norm (nHFEN).

\section{Results and Discussion}

Table 1 summarizes quantitative comparisons across two centers. On the in-house dataset, the proposed SAFE-Diff achieved the best overall performance (SSIM: $0.909 \pm 0.027$, PSNR: $28.246 \pm 4.662$, NMSE: 0.078, nHFEN: 0.819), outperforming MT-DDPM, Cola-Diff, MU-Diff, and the direct $x_0$ regression baseline. Multi-scale attention (baseline with MSA, Multi-Scale Attention) improved SSIM (0.876 → 0.887) and reduced NMSE (0.084 → 0.080), while the full model further improved performance, indicating complementary effects of uncertainty modeling and representation regularization.

\begin{table*}[htbp]
\centering
\caption{Performance comparison across two centers.}
\resizebox{\textwidth}{!}{\begin{threeparttable}
\setlength{\tabcolsep}{6pt}
\begin{tabular}{lcccccccc}
\toprule
\multirow{2}{*}{\textbf{Method}} & 
\multicolumn{4}{c}{\textbf{In-house}} & 
\multicolumn{4}{c}{\textbf{I-SPY2}} \\
\cmidrule(lr){2-5} \cmidrule(lr){6-9}
 & SSIM $\uparrow$ & PSNR $\uparrow$ & NMSE $\downarrow$ & nHFEN $\downarrow$ & SSIM $\uparrow$ & PSNR $\uparrow$ & NMSE $\downarrow$ & nHFEN $\downarrow$ \\
\midrule
MT-DDPM ~\cite{pan20232d} 
& 0.858 $\pm$ 0.045 
& 23.842 $\pm$ 4.324 
& 0.090 $\pm$ 0.046 
& 0.854 $\pm$ 0.203 
& 0.818 $\pm$ 0.067 
& 20.359 $\pm$ 4.791 
& 0.132 $\pm$ 0.086 
& 0.955 $\pm$ 0.386 \\

Cola-Diff ~\cite{jiang2023cola}
& 0.863 $\pm$ 0.040 
& 25.917 $\pm$ 5.639 
& \underline{0.082} $\pm$ 0.035 
& \underline{0.840} $\pm$ 0.197 
& 0.825 $\pm$ 0.057 
& 21.230 $\pm$ 5.322 
& 0.129 $\pm$ 0.099 
& 0.927 $\pm$ 0.376 \\

MU-Diff ~\cite{dayarathna2025mu}
& \underline{0.871} $\pm$ 0.041 
& \underline{26.406} $\pm$ 3.263 
& 0.085 $\pm$ 0.037 
& 0.874 $\pm$ 0.144 
& \underline{0.846} $\pm$ 0.095 
& \underline{23.095} $\pm$ 5.690 
& \underline{0.118} $\pm$ 0.106 
& \underline{0.893} $\pm$ 0.291 \\

\rowcolor{gray!20} 
Baseline 
& 0.876 $\pm$ 0.042 
& 26.569 $\pm$ 3.231 
& 0.084 $\pm$ 0.040 
& 0.899 $\pm$ 0.162 
& 0.849 $\pm$ 0.072 
& 22.817 $\pm$ 4.363 
& 0.117 $\pm$ 0.098 
& 0.921 $\pm$ 0.212 \\

\rowcolor{gray!40} 
Baseline + MSA
& 0.887 $\pm$ 0.043 
& 26.971 $\pm$ 4.320 
& 0.080 $\pm$ 0.038 
& 0.865 $\pm$ 0.205 
& 0.858 $\pm$ 0.068 
& 23.841 $\pm$ 2.163 
& 0.103 $\pm$ 0.089 
& 0.890 $\pm$ 0.292 \\

\rowcolor{gray!50} 
SAFE-Diff
& \textbf{0.909} $\pm$ 0.027
& \textbf{28.246} $\pm$ 4.662
& \textbf{0.078} $\pm$ 0.034
& \textbf{0.819} $\pm$ 0.193
& \textbf{0.870} $\pm$ 0.041
& \textbf{24.537} $\pm$ 4.412
& \textbf{0.091} $\pm$ 0.080
& \textbf{0.882} $\pm$ 0.208 \\
\bottomrule
\end{tabular}
\end{threeparttable}}
\end{table*}

On the external I-SPY2 cohort, consistent trends were observed. The proposed method achieved the highest SSIM ($0.870 \pm 0.041$) and PSNR ($24.537 \pm 4.412$), with the lowest NMSE (0.091). Performance degradation from in-house to I-SPY2 was moderate relative to competing methods, suggesting improved cross-center generalization.

\begin{figure}[htbp]
\centering
\includegraphics[width=\textwidth]{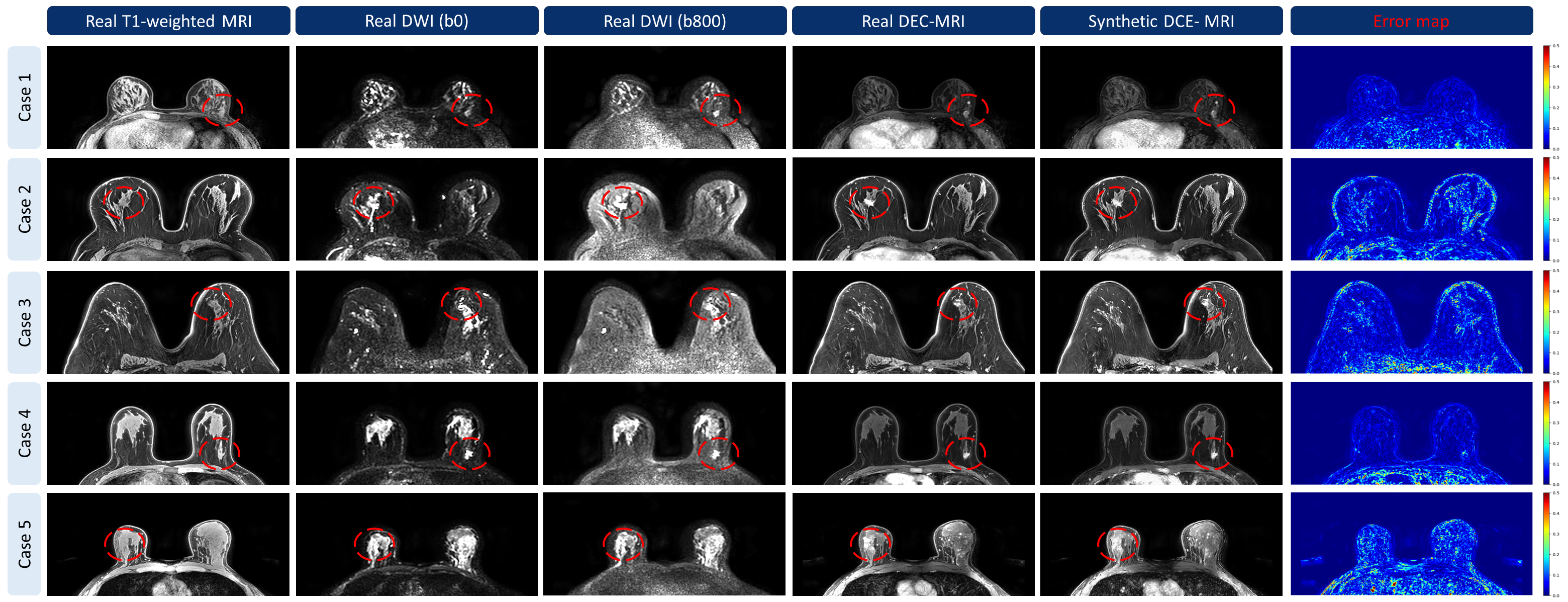}
\caption{Examples for synthetic contrast-enhanced breast MRI.}.
\label{fig2}
\end{figure}

\begin{figure}[htbp]
\centering
\includegraphics[width=\textwidth]
{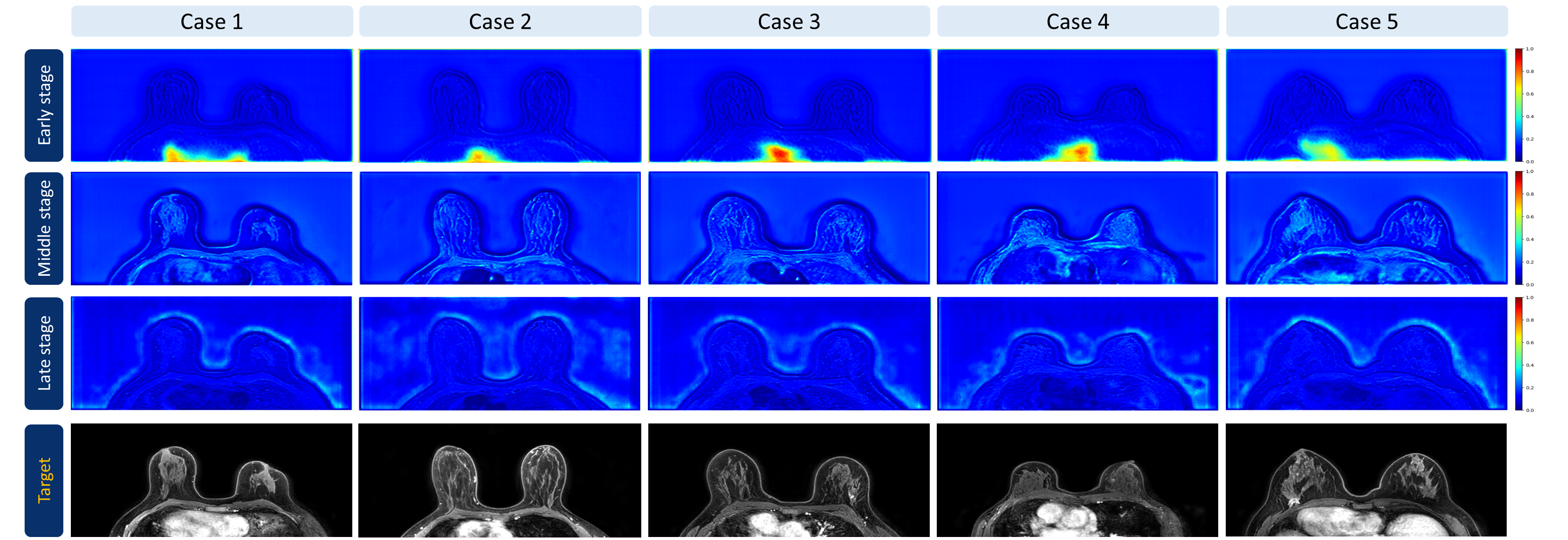}
\caption{Examples for uncertainty maps.}
\label{fig3}
\end{figure}

Qualitative results are shown in Fig.~\ref{fig2}. The synthetic images show comparable lesion location and major morphological features, with observable intralesional enhancement heterogeneity. The error maps indicate that discrepancies are mainly located at anatomical boundaries and other high-frequency structures, while relatively homogeneous areas exhibit lower error. Fig.~\ref{fig3} illustrates the spatial evolution of predicted uncertainty. Early stages show higher uncertainty in thoracic cavity regions, intermediate stages shift uncertainty toward the breast parenchyma, and late stages mainly concentrate uncertainty along the breast contour and other anatomical boundaries, where elevated responses may partially relate to air–tissue interface artifacts. This redistribution is consistent with coarse-to-fine structural refinement during generation.

\begin{table*}[htbp]
\centering
\caption{Ablation study.}
\setlength{\tabcolsep}{5pt}
\resizebox{\textwidth}{!}{\begin{tabular}{lccc cccc cccc}
\toprule
\multirow{2}{*}{\textbf{Method}} 
& \multicolumn{3}{c}{\textbf{Components}} 
& \multicolumn{8}{c}{\textbf{Metrics}} \\
\cmidrule(lr){2-4} \cmidrule(lr){5-12}
& UncA & FDisp & MPer
& SSIM $\uparrow$ & PSNR $\uparrow$ & NMSE $\downarrow$ & nHFEN $\downarrow$ 
& SSIM $\uparrow$ & PSNR $\uparrow$ & NMSE $\downarrow$ & nHFEN $\downarrow$ \\
& & & 
& (Global) & (Global) & (Global) & (Global) 
& (Tumor) & (Tumor) & (Tumor) & (Tumor) \\
\midrule

Baseline 
&  &  &  
& 0.876 $\pm$ 0.042 
& 26.569 $\pm$ 3.231 
& 0.084 $\pm$ 0.040 
& 0.899 $\pm$ 0.162 
& 0.549 $\pm$ 0.227 
& 18.248 $\pm$ 2.660 
& 0.037 $\pm$ 0.038 
& 0.526 $\pm$ 0.112 \\

\rowcolor{gray!25} 
w UncA 
& $\checkmark$ &  &  
& 0.892 $\pm$ 0.038 
& 27.178 $\pm$ 4.590 
& 0.081 $\pm$ 0.038 
& 0.853 $\pm$ 0.204 
& 0.575 $\pm$ 0.184 
& 19.711 $\pm$ 2.349 
& 0.031 $\pm$ 0.019 
& 0.514 $\pm$ 0.092 \\

w UncA \& FDisp 
& $\checkmark$ & $\checkmark$ &  
& 0.897 $\pm$ 0.026 
& 27.486 $\pm$ 3.672 
& 0.080 $\pm$ 0.035 
& 0.829 $\pm$ 0.197 
& 0.613 $\pm$ 0.169 
& 20.318 $\pm$ 2.616 
& 0.029 $\pm$ 0.022 
& 0.499 $\pm$ 0.112 \\

\rowcolor{gray!25} 
SAFE-Diff
& $\checkmark$ & $\checkmark$ & $\checkmark$ 
& \textbf{0.909} $\pm$ 0.027 
& \textbf{28.246} $\pm$ 4.662 
& \textbf{0.078} $\pm$ 0.034
& \textbf{0.819} $\pm$ 0.193
& \textbf{0.655} $\pm$ 0.146
& \textbf{20.537} $\pm$ 2.541
& \textbf{0.027} $\pm$ 0.023
& \textbf{0.495} $\pm$ 0.108 \\
\bottomrule
\end{tabular}}
\end{table*}

Table 2 presents the ablation study. Incorporating uncertainty-aware (UncA) reconstruction increased global SSIM from 0.876 to 0.892 and reduced NMSE from 0.084 to 0.081, with additional gains under tumor-region evaluation. Feature-dispersive (FDis) regularization further improved high-frequency detail preservation, reducing global nHFEN from 0.853 to 0.829 and increasing tumor-region SSIM from 0.575 to 0.613. Integrating the mask-aware perceptual (MPer) loss yielded the strongest overall performance, achieving global SSIM of 0.909 and tumor-region SSIM of 0.655, while further reducing NMSE. Overall, the combined contributions of uncertainty-aware learning, feature-dispersive regularization, and mask-aware perceptual supervision result in improved reconstruction fidelity and cross-domain stability.

\section{Conclusion}
This work presents SAFE-Diff, a multi-scale attention-enhanced diffusion framework for contrast-agent-free breast MRI synthesis, integrating uncertainty-aware reconstruction, feature-dispersive regularization, and mask-aware perceptual supervision. Experiments on a large in-house cohort and an external dataset demonstrate consistent improvements over state-of-the-art diffusion models in structural fidelity, high-frequency detail preservation, and cross-center robustness. The proposed framework further provides pixel-wise uncertainty estimation, supporting interpretable virtual contrast-enhanced imaging. Future work will focus on clinical validation through multi-reader studies and prospective evaluation to assess diagnostic performance and potential clinical utility.

\section*{Acknowledgements}
T.Z and R.M. acknowledge the support from the \textit{SAFE-MRI} project (from Horizon Europe, European Research Council, No 101087701), and the \textit{ODELIA} project (from the European Union’s Horizon Europe research and innovation programme under grant agreement, No 101057091). T.Z. is Editorial Board member of \textit{European Radiology Experimental}, Editorial Board member of \textit{BMC Medicine}, and Trainee Editorial Board member of \textit{Radiology: Artificial Intelligence}. R.M. Mann is advisory Editorial Board member of \textit{European Radiology}, Deputy Editor for \textit{Radiology}, chair of the scientific committee and member of the executive board of the European Society of Breast Imaging, chair of the Dutch College of Breast Imaging. R.M. Mann declares having received grants or research support from the Dutch Research Council, ZonMW, European Research Council, Horizon Europe, Dutch Cancer Society, EFRO Oost, InHolland, Siemens Healthineers, Bayer healthcare, Beckton \& Dickinson, Screenpoint medical, Lunit, Koning, and PA Imaging; consulting fees from Siemens, Bayer, Guerbet, Screenpoint medical, Beckton \& Dickinson; and honoraria for lectures from Siemens, Bayer, and Beckton \& Dickinson.

\section*{Disclosure of Interests}
The authors declare no competing interests.

%
%
%
\bibliographystyle{splncs04}
\bibliography{Paper-3480}

\end{document}